\pdfoutput=1

\documentclass[11pt]{article}

\usepackage[final]{acl}

\usepackage{times}
\usepackage{latexsym}

\usepackage[T1]{fontenc}

\usepackage[utf8]{inputenc}

\usepackage{microtype}

\usepackage{inconsolata}

\usepackage{graphicx}

\usepackage{booktabs}
\usepackage{enumitem}
\usepackage{listings}
\usepackage{placeins}
\usepackage{amsmath}
\usepackage{float}
\usepackage{tabularx}
\usepackage{amssymb}
\newcommand{\yunfan}[1]{}
\newcommand{\smara}[1]{}
\newcommand{\kmnote}[1]{}
\newcommand{\ndnote}[1]{}

\title{Forecasting Conversation Derailments Through Generation}

\author{
 \textbf{Yunfan Zhang\textsuperscript{1}},
 \textbf{Kathleen McKeown\textsuperscript{1}},
 \textbf{Smaranda Muresan\textsuperscript{1,2}}\\
 \textsuperscript{1}Columbia University \quad
 \textsuperscript{2}Barnard College\\
 \texttt{yunfan.z@columbia.edu} \quad
 \texttt{kathy@cs.columbia.edu} \quad
 \texttt{smara@columbia.edu}
}

\begin{document}
\maketitle
\begin{abstract}
Forecasting conversation derailment can be useful in real-world settings such as online content moderation, conflict resolution, and business negotiations. However, despite language models' success at identifying offensive speech present in conversations, they struggle to forecast future conversation derailments. In contrast to prior work that predicts conversation outcomes solely based on the past conversation history, our approach samples multiple future conversation trajectories conditioned on existing conversation history using a fine-tuned LLM. It predicts the conversation outcome based on the consensus of these trajectories. We also experimented with leveraging socio-linguistic attributes, which reflect turn-level conversation dynamics, as guidance when generating future conversations. Our method of future conversation trajectories surpasses state-of-the-art results on English conversation derailment prediction benchmarks and demonstrates significant accuracy gains in ablation studies.

\kmnote{Unlike reviewer 1 (who I might not trust as a good reviewer), I don't feel you have to have the algorithm in the abstract. Could you combine the next two sentences to make it more highlevel. I think what you want in the abstract is what your contribution is. Or, just go back to the original abstract as we discussed.} \yunfan{Sounds good. I will revert to the original abstract.}

\end{abstract}

\section{Introduction}
\label{sec:intro}
\begin{figure}[h!]
\centering%
\includegraphics[width=1\columnwidth]{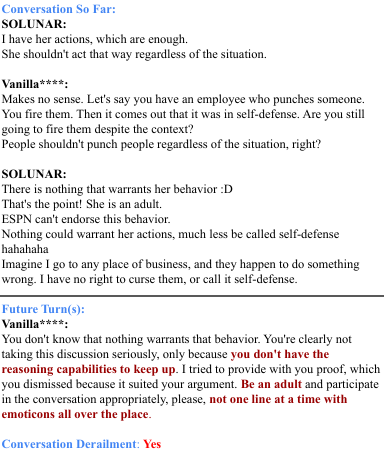}
\caption{
An example conversation 
from the BNC dataset, including background and the future turn. 
Offensive speech is highlighted in red. Our task requires forecasting whether the derailment would occur in the future based on the 
conversation so far. 
}
\label{fig:dataset-example}
\end{figure}

Predicting future derailments from ongoing conversations has a wide range of real-world applications. For instance, in online moderation, the ability to forecast if a discussion thread might devolve into offensive exchanges allows moderators to intervene preemptively. Similarly, in conflict resolution, political hearings, and business negotiations, a system that warns participants of impending conflicts could help mitigate disputes before they escalate.

Despite its utility, predicting future conversation derailments presents significant challenges. As shown in examples in Figure \ref{fig:dataset-example}, unlike the detection of offensive conversational turns~\cite{hate-speech-detection-julia-hirschberg, hate-speech-2, hate-speech-3, hate-speech-4}, which focuses on {\em identifying} harmful speech within a given conversation transcript, our task requires {\em forecasting} whether offensive turns will occur later in the conversation. This makes our task significantly more challenging: the model has to predict potential future derailments based on an otherwise benign conversation history, and this demands a nuanced understanding of the conversation’s progression. Our task is further complicated by the inherently diverse 
nature of human interactions: given the same conversational history, there are multiple possible future trajectories, each with varying topics, styles, and tones. 

To address these challenges, we propose a novel approach that forecasts future derailments by \textit{generating potential continuations of the given conversation}. We define \textit{conversation derailments} as conversations that end with offensive speech or ad hominem attacks, with dataset-specific criteria and annotation procedures explained in Section \ref{ssec:datasets}.
\kmnote{Reviewer 2 was asking for a better definition. If you have one later perhaps point to it here (like "We provide a more formal definition in..." - if in fact you do. } \yunfan{I added a pointer to how CGA-Wiki and BNC determines derailment, which we detailed in section \ref{ssec:datasets}}
Our approach is based on the observation that while Large Language Models (LLMs) may not be effective classifiers for forecasting conversation breakdowns directly, they excel at generating conversation continuations. Thus, we adopt a generate-then-predict approach for forecasting conversation derailments. We first fine-tune an LLM on modeling online conversations. We then sample multiple conversation continuations from this fine-tuned LLM, conditioned on the existing conversation history. We feed each generated continuation along with the existing conversation history to a binary conversation derailment classifier and obtain multiple predictions. We determine the final conversation outcome by taking a majority vote of these individual predictions. By sampling multiple plausible continuations and aggregating the individual predictions with majority vote, we reduce the variability of stochastic LLM outputs, leading to more robust and accurate forecasting of conversation derailments.

Additionally, inspired by prior work \cite{todd-morrill-social-orientation} and the Circumplex Theory \cite{circumplex-1, circumplex-2, circumplex-3} in Psychology, we \textit{explore whether the use of social orientation labels can guide the generation of conversation continuations}. Social orientation labels (e.g., Assertive, Confrontational) are socio-linguistic attributes that reflect conversation dynamics and emotional states (Figure \ref{fig:method-illustration}), and have been shown to help computational models better capture the flow of conversation \cite{todd-morrill-social-orientation}. 

We validate our approach on two datasets: the Conversation Gone Awry Wiki Split (CGA-Wiki) dataset~\cite{conversation-gone-awry, trouble-on-the-horizon}, derived from Wikipedia editor discussions, and the Before Name Calling (BNC) dataset~\cite{before-name-calling}, based on Reddit’s \texttt{r/changemyview} threads. Our method demonstrates significant accuracy improvements compared to the previous state-of-the-art and few-shot GPT-4o. On the CGA-Wiki dataset, we achieve a 4-7\% absolute accuracy improvement, while on the BNC dataset, our method yields an 18-20\% improvement. Extensive ablation studies further confirm the effectiveness of our proposed approach. \kmnote{It's unclear to me... or I don't remember how you get this big improvement on BNC when later you say that performance using BART drops less on BNC than wikipedia (the first set of experiments)} \yunfan{I think this paragraph says we are comparing against previous SotA and GPT-4o, not one of our baselines. We do have 18-20\% improvement over previous SotA and GPT-4o on BNC, so I am keeping this paragraph as is.}

In sum, our contributions are:
\begin{itemize}
    \item{A novel approach for predicting conversation outcomes (derailment or not) by generating multiple potential continuations given the conversation so far.}
    \item{A thorough experimental setup that shows that our proposed approach significantly outperforms prior state-of-the-art methods and powerful new models such as GPT-4o with in-context learning (few-shot, $k=4$).}
    \item{An exploration of whether social orientation labels can guide the generation of conversation continuations, with mixed impact on derailment prediction accuracy, depending on the conversation genre.}
\end{itemize}

Our datasets and models are available through \href{https://github.com/YunfanZhang42/ConversationDerailments}{this GitHub repository}.

\section{Problem Statement and Motivation}
\label{sec: problem_statement_and_motivation}

Our objective is to estimate the likelihood of future conversation derailment based on the benign conversation history up to a certain point. We define conversation derailments as offensive speech and \textit{ad hominem} attacks, in line with previous work \cite{conversation-gone-awry, before-name-calling}.

Formally, let $C = \{c_1, \dots, c_n\}$ represent the conversation history consisting of $n$ turns, where $c_i$ denotes the $i$-th turn in the dialogue. Let $y(\{c_1, \dots, c_i\})$ be a binary indicator function for conversation derailment over the turns $\{c_1, \dots, c_i\}$, where $y=1$ indicates the presence of derailments.
\yunfan{Based on the expression below, I think $y=1$ does only refer to the presence of derailment in the current turns, not future turns. We want to predict future derailments from current benign turns, so the goal is to predict $P(y(\{c_{k+1}, \dots, c_n\}) = 1)$ when we have $y(\{c_1, \dots, c_k\}) = 0, k<n$}

Our goal is to estimate the likelihood of future derailments given the first $k$ benign turns, expressed as:
\begin{align*}
P(y(\{c_{k+1}, \dots, c_n\}) = 1 \mid \{c_1, \dots, c_k\}) \\
\text{subject to } k < n, y(\{c_1, \dots, c_k\}) = 0
\end{align*}

\begin{table*}[h]
\fontsize{9}{9}\selectfont
\renewcommand{\arraystretch}{1.2}
\centering
\begin{tabular}{l|llll|llll}
\toprule
\multicolumn{1}{c}{} & \multicolumn{4}{c}{CGA-Wiki} & \multicolumn{4}{c}{BNC} \\
\cmidrule(r){2-5} \cmidrule(r){6-9}
Methods & Acc. & Prec. & Rec. & F1 & Acc. & Prec. & Rec. & F1 \\
\midrule
BART, No Offensive Turns in Input & 65.4 & 63.9 & 70.5 & 67.0 & 84.2 & 85.6 & 82.3 & 83.9 \\
BART, All Turns & \textbf{95.5} & \textbf{94.2} & \textbf{96.9} & \textbf{95.5} & \textbf{92.7} & \textbf{91.7} & \textbf{93.8} & \textbf{92.8} \\
\bottomrule
\end{tabular}
\caption{\label{tab:motivation} Accuracy, precision, recall, and F1 scores on \textbf{CGA-Wiki} and \textbf{BNC}. Modern Language Models (LMs) can easily identify offensive speech present in the conversation. However, we notice a significant drop in accuracy when only the benign speech is given and the model is trained to forecast offensive speech in future exchanges.}
\end{table*}

To illustrate the challenges of forecasting conversation derailment, we conducted an experiment comparing the performance of a language model in two settings: detecting offensive speech from a full conversation transcript and predicting future derailments based only on the conversation history prior to any offensive speech. We used BART-Base \cite{bart} as the underlying model, trained with a Binary Cross Entropy loss to directly predict the conversation derailment.

The results, shown in Table \ref{tab:motivation}, demonstrate that even smaller LMs such as BART perform well in detecting offensive speech when given full transcripts including the offensive turns, achieving over 90\% accuracy on both the CGA-Wiki and BNC datasets. However, the model's performance drops substantially when forecasting future derailments from only benign turns. Specifically, BART's accuracy falls to 65\% on CGA-Wiki and 84\% on BNC, representing absolute reductions of 30\% and 8\%, respectively.

These findings emphasize a key challenge in this task: while LMs excel at identifying offensive speech present in the transcript, they struggle significantly when predicting upcoming derailments based solely on benign conversation history. This performance gap suggests that, in the absence of overt offensive speech, LMs have difficulty discerning the subtle conversational cues that may indicate a future shift toward derailments.

\section{Methodology}
\label{sec:methodology}


\begin{figure*}[h!]
\centering%
\includegraphics[width=\textwidth]{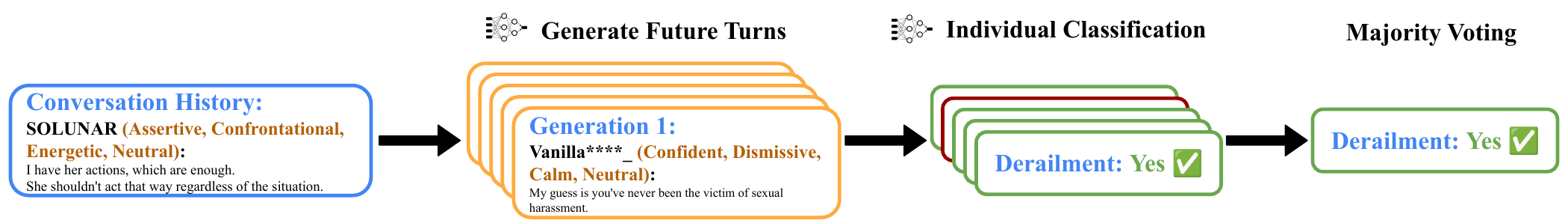}
\caption{An illustration of our methodology. Social orientation labels are highlighted in brown. 
We sample multiple potential conversation continuations from a given conversation history. Then, we predict individual conversation outcomes by combining each continuation with the given conversation history. We use the majority of the individual results to predict our final conversation outcome.}
\label{fig:method-illustration}
\end{figure*}

Figure \ref{fig:method-illustration} shows an overview of our approach. We employ a fine-tuned LLM to generate multiple future continuations given the existing conversation history. We then train a dedicated classifier to assess the conversation outcome by analyzing both the existing and newly generated conversations. We determine the final outcome based on the majority vote of these individual predictions. We also study whether social orientation labels (see Section \ref{ssec:social_orientation_labels}) can be used to guide the text generation and conversation derailment classification.

\subsection{Fine-tuning Conversation LLMs}
\label{ssec:fine-tuning-conversation-generation-llms}
Although it is possible to generate future conversations directly with LLMs through only few-shot prompting, we discovered that LLMs are unlikely to generate offensive comments out of the box due to their built-in safety mechanism. We therefore fine-tune LLMs on conversation transcripts from the CGA-Wiki and BNC datasets, enhancing the models' capability to produce authentic-sounding comments that are contextually aligned with the existing conversation history. We also experimented with prepending social orientation labels (further explained in Section \ref{ssec:social_orientation_labels}) to each comment to steer the text-generation process. 

Formally, consider an LLM $f$ with parameters $\psi$. Let $D = \{C_1, \dots, C_n\}$ be a dataset of conversation transcripts, where each conversation $C_j \in D$ consists of turns $C_j = \{c_1, \dots, c_n\}$. Each turn $c_i$ may optionally be associated with a social orientation label $s_i$, forming a corresponding set of labels $S_j = \{s_1, \dots, s_n\}$.

During training, the first $k$ turns and the optional social orientation labels $\{(s_1, c_1), \dots, (s_k, c_k)\}$ are used as the input context, denoted as $\mathbf{x}_j$. The model is trained to jointly predict the subsequent sequence of social orientation labels (if used) and conversation turns $\{(s_{k+1}, c_{k+1}), \dots, (s_n, c_n)\}$, which forms $\mathbf{y}_j$. The training objective is: 
\[
    \min_{\psi} \mathcal{L}(\psi) = - \sum_{j=1}^{\left| D \right|} \sum_{t=1}^{\left| \mathbf{y}_j \right|} \log p\left(\mathbf{y}_{j,t} \mid \mathbf{y}_{j,<t}, \mathbf{x}_j; f_\psi\right)
\]
\subsection{Training Derailment Classifiers}
\label{ssec:fine-tuning-conversation-outcome-classifiers}


We train a separate classifier to predict the likelihood of future conversation derailments by leveraging both the existing conversation history and the generated future turns. To ensure our classifier can generalize to synthetic conversation turns, we augment the training dataset with hypothetical future conversations generated by our fine-tuned conversation generation model. 

Formally, for each conversation $C_j = \{c_1, \dots, c_n\}$, $C_j \in D$, and a corresponding set of social orientation labels $S_j$, we use our fine-tuned LLM $f_\psi$ to sample $l$ potential future continuations, denoted as below. 
\begin{align*}
& \{(s_{k+1}^i, c_{k+1}^i),\ \dots,\ (s_{n}^i, c_{n}^i)\}_{i = 1}^{l} \\
& = f_\psi\big(\{(s_1, c_1),\ \dots,\ (s_k, c_k)\}\big)
\end{align*}

We experiment with both configurations: one where the social orientation labels are included and one where they are excluded. We then train the conversation derailment classifier, $f_\phi$, using both the synthetic future conversation turns and the real future conversation turns from the original dataset, supervised by the ground truth label given by the original dataset. Namely, we have:
\begin{align*}
\min_{\phi} \mathcal{L}(\phi) &= - \sum_{j=1}^{m} \Big[ y_j \log f_\phi(X_j) \\
&\quad + (1 - y_j) \log \left(1 - f_\phi(X_j)\right) \Big]
\end{align*}

Where $X_j$ represents the combined sequence of existing conversation turns and either real or synthetically generated future turns, and $y_j$ is the ground truth label indicating the presence ($y_j = 1$) or absence ($y_j = 0$) of conversation derailment.


\subsection{Inference Time Majority Voting}
\label{ssec:test-time-methodology}
At inference time, given a conversation history, we generate $L$ hypothetical conversation continuations using our fine-tuned conversation generation LLM $f_\psi$. Each generated continuation is appended to the existing conversation history and then classified by conversation outcome classifier $f_\phi$, resulting in $L$ individual outcome predictions. The final conversation outcome is determined by taking the majority vote across these $L$ predictions. This procedure compensates for the inherent randomness in LLM decoding, thereby reducing the influence of outlier generations and improving the robustness and accuracy of the final prediction.



\subsection{Social Orientation Labels}
\label{ssec:social_orientation_labels}

We explore the use of social orientation labels in modeling conversation dynamics and examine their helpfulness on predicting conversation outcomes. Our social orientation labels are inspired by Circumplex Theory \cite{circumplex-1, circumplex-2, circumplex-3} in Psychology, which characterizes interpersonal interactions by assigning descriptive labels along a set of core dimensions. Figure \ref{fig:method-illustration} provides an example of our social orientation analysis. Our social orientation axes are defined as follows:

\begin{description}[align=left, leftmargin=0pt, style=unboxed] 
\item[Power] \cite{circumplex-1} captures the extent to which an individual seeks to control or assert dominance in the conversation. The available labels for this dimensions are \texttt{assertive, confident, neutral, open-minded, submissive}.

\item[Benevolence] \cite{circumplex-1} reflects the warmth and positivity of the interaction. The available labels are \texttt{confrontational, dismissive, neutral, friendly, supportive}.

\item[Arousal] \cite{circumplex-2} indicates the level of energy or excitement expressed in the comment. The available labels are \texttt{energetic, neutral, calm}.

\item[Political Leaning] \cite{circumplex-3} assesses the political inclination conveyed by the comment. The available labels are \texttt{liberal, neutral, conservative}.
\end{description}

Building on the methodology outlined in \citet{todd-morrill-social-orientation}, we annotate a set of social orientation labels on a turn-by-turn basis using GPT-4o. We extend \citet{todd-morrill-social-orientation} by decoupling the axes and prompting GPT-4o to consider labels for each axis independently. Each turn is assigned one label from each of the four psychological dimensions. Formally, given an LLM $f_\theta$, a few-shot prompt $p$, and a conversation $C = \{c_1, \dots, c_n\}$, the corresponding set of social orientation labels $S = \{s_1, \dots, s_n\}$ is computed as follows:
\[\{s_1, \dots, s_n\} = f_\theta(p, \{c_1, \dots, c_n\})\]

Each $s_i \in S$ consists of one keyword from each of the four axes. For instance, an example $s_i$ could be \texttt{assertive, confrontational, energetic, conservative}.


\section{Experiments}
\label{sec:experiment-setup}
We evaluate our conversation derailment prediction method on two datasets: Conversation Gone Awry Wiki Split \cite{conversation-gone-awry, trouble-on-the-horizon} and the Before Name Calling dataset \cite{before-name-calling}. We describe the datasets and our experiment details below.

\subsection{Datasets}
\label{ssec:datasets}

\begin{description}[align=left, leftmargin=0pt, style=unboxed]
\item[Conversation Gone Awry Wiki Split (CGA-Wiki)] is a conversation derailment dataset derived from the discussion history between Wikipedia editors. In this dataset, derailments are defined as personal attacks or ad hominem speech. These labels were provided by paid human annotators. We use both the original dataset \cite{conversation-gone-awry} and the additional samples annotated in \cite{trouble-on-the-horizon}, resulting in a total of 4,188 samples. After excluding section headers without actual conversation turns, each sample comprises 3 to 19 turns, with a median of 6 turns. The first $n-1$ turns in each sample are benign, while the final turn is either benign or offensive with equal probability.

Following prior work \cite{cga-wiki-sota-1, cga-wiki-sota-2, cga-wiki-sota-3, todd-morrill-social-orientation}, for the primary results in Section \ref{ssec:performance_cga_wiki} and Table \ref{tab:performance_cga_wiki}, we treat the first $k = n-1$ turns as the conversation history and input to our model. We then predict whether the last turn will be benign or offensive.


To assess the impact of generating multiple future turns, we experiment with models given limited conversation history ($k=2$ or $k=4$ turns). Since the remaining number of turns before the conversation ends could be considered future information, we do not enforce a fixed number of generated turns. Instead, we allow the model to continue until it deems the conversation complete. This setup enables us to analyze how extending predictions further into the future affects performance. The results are presented in Section \ref{ssec:fewer_input_turns} and Table \ref{tab:performance_cga_wiki_turns_ablation}.

We adhere to the official training, validation, and test splits, consistent with \citet{trouble-on-the-horizon}.

\yunfan{Do we need to cite them all here? Seems verbose.}

\item[Before Name Calling (BNC)] \cite{before-name-calling} is a conversation derailment dataset sourced from posts and replies in the Reddit \texttt{r/changemyview} community, containing 2,582 samples. Each sample includes $n=4$ consecutive conversation turns from the same comment-reply thread. The first $k=3$ turns are benign, while the 4th turn represents either a constructive or derailing outcome: a constructive outcome is one where the reply successfully changes the original poster’s view, as evidenced by an upvote from the OP; a derailing outcome is one where the reply is flagged by moderators for containing \textit{ad hominem} or otherwise offensive content. These two outcomes are represented with equal probability in the dataset. We are aware that \citet{conversation-gone-awry} also compiled a conversation derailment dataset (CGA-CMV) using comments from Reddit \texttt{r/changemyview} community; however, we prefer BNC over CGA-CMV as the BNC dataset contains fewer false-negatives upon our manual examination.

As in \citet{before-name-calling}, we treat the first 3 turns as conversation history and used them as inputs to our model. We then predict whether the 4\textsuperscript{th} turn would be benign or contain \textit{ad hominem} attacks. Because all samples in this dataset are limited to only 4 turns, we do not experiment with generating multiple future turns on this dataset. Since no official split is provided for this dataset, we randomly divide it into training, validation, and test sets with an 8:1:1 ratio.
\end{description}

\subsection{Experiment Setup}
\label{ssec:training-and-evaluation}
\begin{description}[align=left, leftmargin=0pt, style=unboxed]
\item[Annotating Social Orientation Labels.] We use GPT-4o (gpt-4o-2024-05-13) \cite{gpt-4} with few-shot prompting ($k=4$) to annotate the social orientation labels. The prompt used in this process is provided in Appendix \ref{text:social-orientation-annotation-template}.

\item[Fine-tuning Conversation Generation LLMs.] We fine-tune the Mistral-7B-Base model \cite{mistral7b} following the procedures outlined in Section \ref{ssec:fine-tuning-conversation-generation-llms}. We use Low Rank Adaptation (LoRA) \cite{lora} to conserve GPU memory and make training feasible on our hardware. Further details about our fine-tuning setup are available in Table \ref{tab:gen-llm-hyper-params} in Appendix \ref{ssec:model_training_evaluation_details}.

\item[Training Conversation Outcome Classifiers.] After fine-tuning Mistral-7B for conversation generation, we augment our training set with hypothetical future turns generated by our fine-tuned model, as described in Section \ref{ssec:fine-tuning-conversation-outcome-classifiers}. We generate $l=2$ hypothetical continuations for each training sample, and then fine-tune both a BART-Base model and a Mistral-7B-Base model to classify whether a derailment occurs in the transcript. We use the two synthetic conversation continuations along with the real future conversation turns as the input, and the gold derailment labels provided by the datasets as the target label in fine-tuning. The training hyper-parameters for the BART classifier are provided in Table \ref{tab:bart-classifier-hyper-params} in Appendix \ref{ssec:model_training_evaluation_details}, and those for the Mistral classifier are in Table \ref{tab:mistral-classifier-hyper-params}.

\item[Inference Time Configurations.] We follow the inference strategy outlined in Section~\ref{ssec:test-time-methodology}. To determine $L$, the number of continuations per sample, we conduct an ablation study on the CGA-Wiki and BNC validation sets, as detailed in Appendix~\ref{ssec:performance_continuations_ablation} and Table~\ref{tab:performance_cga_wiki_continuations_ablation}. We vary $L$ from 1 to 15 and observe that increasing $L$ from 1 to 5 yields a 1-2\% improvement in accuracy. Beyond $L=5$, gains are marginal (0.1-0.4\%). Based on this, we fix $L=5$ for all experiments to strike a balance between prediction accuracy and computational overhead. 

\item[Baseline Settings.] For both the CGA-Wiki and BNC datasets, we train a BART-Base classifier and a Mistral-7B classifier as our baseline models. The baseline models are trained to predict the gold labels given by the dataset using the first $k$ turns as input. We do not incorporate intermediate steps such as social orientation labels or future conversation turn generation.

\item[Metrics.] We evaluate the performance of our approaches on accuracy, precision, recall, and F-1. We consider the presence of offensive or \textit{ad hominem} speech as the positive case for the purpose of calculating precision and recall. 
\end{description}

\section{Results and Analysis}
\label{sec:results}



\label{ssec:performance_cga_wiki}
\begin{table*}[h]
\fontsize{9}{9}\selectfont
\renewcommand{\arraystretch}{1.2}
\centering
\begin{tabular}{l|l|llll|llll}
\toprule
\multicolumn{1}{c}{} & \multicolumn{1}{c}{} & \multicolumn{4}{c}{CGA-Wiki} & \multicolumn{4}{c}{BNC} \\
\cmidrule(r){3-6} \cmidrule(r){7-10}
Methods & Category & Acc. & Prec. & Rec. & F1 & Acc. & Prec. & Rec. & F1 \\
\midrule
\citet{todd-morrill-social-orientation} & SotA & 65.5 & - & - & - & - & - & - & - \\
\citet{cga-wiki-sota-1} & SotA & 66.9 & 63.3 & 80.2 & 70.8 & - & - & - & - \\
\citet{cga-wiki-sota-2} & SotA & 64.3 & 61.2 & 78.9 & 68.8 & - & - & - & - \\
\citet{cga-wiki-sota-3} & SotA & 65.2 & 64.2 & 69.1 & 66.5 & - & - & - & - \\
\citet{cga-wiki-sota-4} & SotA & 67.1 & 67.1 & 66.9 & 66.9 \\
\citet{before-name-calling} & SotA & - & - & - & - & 72.1 & - & - & - \\
GPT-4o Few-shot & SotA & 65.1 & 60.3 & \textbf{88.8} & 71.8 & 75.7 & 68.7 & 94.6 & 79.6 \\
\midrule
BART SFT & Baseline & 65.4 & 63.9 & 70.5 & 67.0 & 84.2 \phantom{$\star$} $\bullet$ & 85.6 & 82.3 & 83.9 \\
BART + SO & Ablation & 63.6 & 64.2 & 61.4 & 62.8 & 87.6 \phantom{$\star$} $\bullet$ & 90.8 & 83.9 & 87.2 \\
BART + G & Proposed & 69.2 $\star$ & 67.5 & 73.8 & 70.5 & 89.2 $\star$ $\bullet$ & 91.1 & 86.9 & 88.9 \\
BART + SO + G & Proposed & 69.2 $\star$ & 67.1 & 75.2 & 70.9 & 91.1 $\star$ $\bullet$ & 89.6 & 93.1 & 91.3 \\
\midrule
Mistral SFT & Baseline & 64.5 & 60.5 & 83.8 & 70.3 & 90.4 \phantom{$\star$} $\bullet$ & 92.0 & 88.5 & 90.2 \\
Mistral + SO & Ablation & 64.9 & 62.4 & 75.0 & 68.1 & 89.6 \phantom{$\star$} $\bullet$ & 89.9 & 89.2 & 89.6 \\
Mistral + G & Proposed & \textbf{71.4} $\star$ $\bullet$ & \textbf{68.4} & 79.5 & \textbf{73.6} & 93.1 \phantom{$\star$} $\bullet$ & \textbf{95.2} & 90.8 & 92.9 \\
Mistral + SO + G & Proposed & 68.6 $\star$ & 68.1 & 69.8 & 68.9 & \textbf{93.8} $\star$ $\bullet$ & 93.2 & \textbf{94.6} & \textbf{93.9} \\
\bottomrule
\end{tabular}

\caption{\label{tab:performance_cga_wiki} Accuracy, precision, recall, and F1 scores on \textbf{CGA-Wiki} and \textbf{BNC}. 
SFT stands for supervised fine-tuning, SO stands for social orientation labels, and G stands for generation.
$\star$ denotes statistically significant (z-test $p<0.1$) accuracy gains compared to the baselines. 
$\bullet$ denotes statistically significant (z-test $p<0.1$) accuracy gains compared to the best-performing state of the art.
On both datasets, methods that leverage conversation generation consistently outperforms the state-of-the-art methods and the baselines. Social orientation labels further improve performance on BNC when paired with conversation generation.}
\end{table*}

We show our experiment results on Conversation Gone Awry Wiki Split (CGA-Wiki) and Before Name Calling (BNC) in Table \ref{tab:performance_cga_wiki}. 

\subsection{Comparisons with the State of the Art}
\label{ssec:improvement_over_sota}
Our prediction-through-generation approach achieves significant improvements over previous state-of-the-art fine-tuned models and commercial LLMs such as GPT-4o on both CGA-Wiki and BNC datasets.

As shown in Table \ref{tab:performance_cga_wiki}, on the CGA-Wiki dataset, our best-performing model, the Mistral classifier with conversation generation (Mistral + G), surpasses the best previous fine-tuned models in accuracy, precision, and F1 score, while slightly trailing \citet{cga-wiki-sota-2} in recall. It also outperforms GPT-4o few-shot in accuracy, precision, and F1 score. While our model falls short of GPT-4o in recall, GPT-4o’s high recall is artificially inflated due to its tendency to over-predict derailment. Despite balanced class labels (50\%) in the datasets and few-shot examples, GPT-4o classifies 73.6\% of conversations as derailments, leading to an inflated recall. 
\smara{I left only the overall metrics, and cut discussion on recall as i felt is not needed} \yunfan{I think some of the reviewers asked about GPT-4o recall. I have put the statements back.}

Similarly, on the BNC dataset, we also see improvements with our prediction-through-generation approach. Our best variant, the Mistral classifier with conversation generation and social orientation labels (Mistral + SO + G) significantly outperforms the previous state-of-the-art \cite{before-name-calling} by a margin of 21.7 in absolute accuracy. It also significantly exceeds GPT-4o few-shot in accuracy, precision, and F1 score, while matching recall. Mistral + G variant also achieves similar levels of improvements over the previous state-of-the-art and GPT-4o in accuracy, precision, and F1. These results underscore the effectiveness of our approach over prior work and powerful commercial LLMs such as GPT-4o.

\subsection{Impact of Prediction-through-generation}
\label{ssec:impact-of-prediction-through-generation}
Our ablation studies demonstrate the effectiveness of the prediction-through-generation approach on both the CGA-Wiki and BNC datasets. As shown in Table \ref{tab:performance_cga_wiki}, models incorporating prediction-through-generation consistently outperform their baseline counterparts, which are fine-tuned directly on gold labels from the dataset. On the CGA-Wiki dataset, both BART + G and Mistral + G achieve higher accuracy, precision, and F1 scores compared to their respective baselines. Additionally, BART + G also improves recall over the baseline BART, while Mistral + G experiences a slight decline in recall. On the BNC dataset, both BART + G and Mistral + G surpass their baselines across all evaluation metrics, including accuracy, precision, recall, and F1 score.

\subsection{Impact of Social Orientation Labels}
\label{ssec:impact-of-social-orientation-labels}
Inspired by previous work \cite{todd-morrill-social-orientation}, we explored adding social orientation labels to our conversation derailment pipeline. To contextualize our approach, we included a worked example of social orientation labels from the BNC dataset in Appendix \ref{ssec:gen_examples}.

As in Table \ref{tab:performance_cga_wiki}, we found mixed results on the effect of social orientation labels, depending on the dataset. The addition of social orientation labels enhanced the prediction of conversation derailment on the BNC dataset. For both Mistral + G + SO and BART + G + SO, adding social orientation labels on top of conversation generation improved accuracy, recall, and F1 score, though there is a slight decrease in precision. However, incorporating social orientation labels did not enhance conversation derailment detection accuracy on CGA-Wiki dataset. 

To better understand the mixed contribution of social orientation labels, we conducted a human evaluation of the GPT-4o-annotated labels, as described in Appendix \ref{ssec:so_human_eval}. Our results indicate that while GPT-4o achieves a reasonable annotation accuracy, with 70\% agreement with human annotators, this accuracy remains lower than our conversation outcome prediction performance (71.4\% on CGA-Wiki and 93.8\% on BNC). This suggests that the social orientation labels may still be too noisy, providing limited additional information and thereby constraining their overall usefulness in conversation outcome prediction.

\subsection{Impact of Generating Multiple Future Turns}
\label{ssec:fewer_input_turns}
\begin{table*}[h]
\fontsize{9}{9}\selectfont
\renewcommand{\arraystretch}{1.2}
\centering
\begin{tabular}{l|l|llll|llll}
\toprule
\multicolumn{1}{c}{} & \multicolumn{1}{c}{} & \multicolumn{4}{c}{First 2 Turns} & \multicolumn{4}{c}{First 4 Turns} \\
\cmidrule(r){3-6} \cmidrule(r){7-10}
Methods & Category & Acc. & Prec. & Rec. & F1 & Acc. & Prec. & Rec. & F1 \\
\midrule
BART SFT & Baseline & 55.6 & 57.1 & 45.0 & 50.3 & 64.1 & 62.8 & 69.1 & 65.8 \\
BART + G & Proposed & \textbf{56.4} & \textbf{58.2} & 45.5 & 51.1 & \textbf{64.9} & 63.0 & \textbf{71.9} & \textbf{67.2} \\
\midrule
Mistral SFT & Baseline & 56.0 & 56.2 & 54.3 & 55.2 & 61.9 & 61.5 & 63.6 & 62.5 \\
Mistral + G & Proposed & 54.5 & 53.4 & \textbf{70.2} & \textbf{60.7} & 62.5 & \textbf{63.6} & 58.3 & 60.9 \\
\bottomrule
\end{tabular}
\caption{\label{tab:performance_cga_wiki_turns_ablation} Accuracy, precision, recall, and F1 scores on \textbf{CGA-Wiki}, but with only the first 2 or 4 turns as inputs to the model. SFT stands for supervised fine-tuning, and G stands for conversation generation. BART classifier with future conversation generation (BART + G) consistently achieves the highest accuracy in both first 2 turns and first 4 turns settings.}
\end{table*}

We investigate the impact of generating multiple future turns on conversation derailment prediction. Instead of stipulating the number of future turns to generate, which could be considered leaking future information, we restrict the input context to the first $k=2$ and first $k=4$ turns, and then allow the model to generate multiple subsequent turns until it determines the conversation is complete. We observe that with the first 2 turns, the model generates a median of 3 more turns, while with the first 4 turns, it generates a median of 2.

As shown in Table \ref{tab:performance_cga_wiki_turns_ablation}, performance declines in both the "first 2 turns" and "first 4 turns" scenarios compared to the full $n-1$ turn input. This result is expected, as the model operates with reduced conversational context and needs to predict further into the future. However, in both cases, the method employing the BART classifier with future conversation generation (BART + G) maintained the highest accuracy. While the improvement is more modest (approximately 1\%) compared to the full $n-1$ turn setting, the prediction-through-generation approach still proves beneficial.

\subsection{Diversity of Multiple Conversation Continuations}
\label{ssec:diversity-continuations} 

We calculate the BLEU score between different continuations given the same conversation history to evaluate the diversity of generated conversation continuations. More specifically, for each set of $L = 5$ continuations generated for the same conversation, we compute BLEU score by treating one continuation as the hypothesis and the remaining four as references. We repeat this for all possible (4+1) combinations and report the average BLEU score. Lower scores indicate higher diversity among the generated continuations. We focus on the top-performing methods on each dataset: Mistral + G on CGA-Wiki and Mistral + SO + G on BNC.

We found our technique yields highly diverse conversation continuations, with average BLEU scores between inputs at 0.034 for Mistral + G on CGA-Wiki and 0.046 for Mistral + SO + G on BNC. These low scores indicate that the generated continuations are substantially dissimilar from each other, allowing our method to forecast a broad and varied set of plausible conversational trajectories.

\section{Related Work}
\label{sec:related-work}
\subsection{Offensive Speech Detection}
\label{ssec:communication_change_prediction_vs_toxic_speech}

Offensive speech detection is a well-established research area in NLP \cite{hate-speech-detection-julia-hirschberg, hate-speech-2, hate-speech-3, hate-speech-4}. It aims to identify offensive content already present in a single conversation turn or full conversation. 

In contrast, we focus on forecasting potential future derailments, such as offensive speech, without any direct evidence of harmful language in the current conversation. This presents a greater challenge, as the prediction must rely solely on the benign portion of the conversation and infer whether the dialogue might devolve in the future.

\subsection{Predicting Conversation Derailment}
\kmnote{You are mixing present and past tense. Use one consistently throughout} \yunfan{I have change the tense to present.}
\label{ssec:related-work-predict-communication-change}
Several studies address the challenge of predicting conversation derailment using existing conversation history. \citet{conversation-gone-awry, trouble-on-the-horizon} create the CGA dataset by manually identifying \textit{ad hominem} comments within Wikipedia editor discussions. \citet{before-name-calling} compile BNC dataset by using moderation results and upvotes from comments in Reddit \texttt{r/changemyview} community. \citet{cga-wiki-sota-2} propose adopting dynamic forecast window and pretrained LMs to predict derailments on CGA. \citet{cga-wiki-sota-3} apply a hierarchical transformer-based framework on CGA. \citet{todd-morrill-social-orientation} improve performance on the CGA by utilizing turn-level social orientation labels annotated by GPT-4. \citet{summarizing-conversation-dynamics} suggest employing conversation-level natural language summaries generated by GPT-4 to improve conversation derailment prediction performance. \citet{cga-wiki-sota-1} propose using graph neural networks for predicting derailments on CGA. \citet{cga-wiki-sota-4} explore using LLMs such as GPT-4 and Llama 3 to create synthetic training data for CGA. 

In contrast to these methods, which rely on existing conversation history only, our approach generates plausible future turns and bases its prediction on both the existing dialogue and these potential continuations. This allows our derailment classifier to consider how the conversation might unfold in its decision.

\smara{In the other sections you have contrast statements, and here you do not. I think is good to have. I see kathy had a note on this earlier} \yunfan{I added a short contrast.}

\subsection{Social Orientation Labels for Understanding Conversation Dynamics}
\label{ssec:related-work-social-orientation}
Social orientation labels are proposed to analyze the dynamics in conversations \cite{todd-morrill-social-orientation}. Social orientation is a concept originally developed from Circumplex Theory \cite{circumplex-1, circumplex-2, circumplex-3} in Psychology. 
The Circumplex Model allows for the analysis of interpersonal interactions along defined axes like power, benevolence, arousal, and political leaning, each with contrasting traits (e.g. assertive vs. submissive for power). In our proposed approach, we adopt social orientation annotations to guide our LLM in generating future exchanges.

\section{Conclusion and Future Work}
\label{sec:conclusion}


We introduced a novel approach for forecasting conversation derailments by generating potential future conversation trajectories based on existing conversation history. This technique demonstrated strong performance on conversation derailment prediction benchmarks, such as the CGA-Wiki and BNC datasets. We validated the effectiveness of our method through comparisons with state-of-the-art models and comprehensive ablation studies. We also assessed the contribution of social orientation labels in guiding derailment prediction. Future research could extend our methodology to conversations beyond online discourse, such as in-person conversations and meetings.


\section{Limitations}
\label{limitations}
Our work has several limitations that may be addressed by future work. The primary limitation is the scope of conversation domains in our experiments. Due to the limited availability of derailment prediction datasets, we restricted our experiments to the CGA-Wiki dataset, derived from Wikipedia editor discussions, and the BNC dataset, based on the Reddit \texttt{r/changemyview} community. Future studies could explore derailment prediction using datasets from other online sources, as well as in-person conversations.

Additionally, our prediction-through-generation approach has room for improvement. A common failure in our method is generating turns for the wrong speaker. This leads to inaccuracies in tone and content because the model lacks information about which speaker will participate in the next turn. While it is possible to prompt the model with the correct speaker's name, we chose not to do so in order to maintain parity with prior work. Future research could explore whether specifying the next speaker improves performance, especially in scenarios where knowing the next speaker is a reasonable assumption.

Lastly, our method is significantly more computationally intensive compared to previous approaches. We used approximately 1,000 GPU hours for all experiments, with the majority of time spent on fine-tuning and inference of large language models. Future work could focus on enhancing computational efficiency without compromising accuracy.

\section{Ethical Considerations}
\label{ethical_considerations}
For our study on predicting conversation derailment, we used two publicly available datasets: CGA-Wiki, licensed under the MIT license, and BNC, licensed under Apache 2.0. We believe our use of these datasets complies with fair-use guidelines. Neither dataset contains personally identifiable information. However, because these datasets are intended to identify conversational derailments, they inevitably include offensive language.

While we do not anticipate significant risks arising from our work, we acknowledge that our methodology could be applied to online moderation, raising potential concerns about censorship if misused.

\section{Supplementary Materials Availability Statement}
Our code, dataset, and model weights are available at this GitHub repository: \url{https://github.com/YunfanZhang42/ConversationDerailments}.

\section{Acknowledgments}
This research is being developed with funding from the Defense Advanced Research Projects Agency (DARPA) Cross-Cultural Understanding program under Contract No HR001122C0034. The views, opinions and/or findings expressed are those of the authors and should not be interpreted as representing the official views or policies of the Department of Defense or the U.S. Government.

\FloatBarrier

\bibliography{custom}

\clearpage

\appendix

\section{Appendix}
\label{sec:appendix}
\subsection{Conversation Generation Examples}
\label{ssec:gen_examples}

\begin{figure}[h]
\centering%
\includegraphics[width=1\columnwidth]{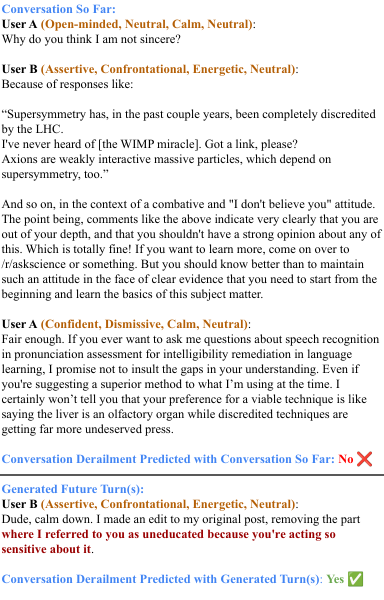}
\caption{
An example conversation from the BNC dataset, including background and the future turn as generated by our fine-tuned LLM. Social orientation labels are highlighted in brown. Offensive speech is highlighted in red. When only given the benign conversation history, the classifier fails to forecast if derailments would happen in the future. Generating the future conversation turns and providing the future turns to the classifier allows the classifier to forecast derailments correctly.
}
\label{fig:example-1}
\end{figure}

\subsection{Ablation Studies on Number of Continuations}
\label{ssec:performance_continuations_ablation}
 
\begin{table*}[h!]
\fontsize{9}{9}\selectfont
\renewcommand{\arraystretch}{1.2}
\centering
\begin{tabular}{l|llll|llll}
\toprule
\multicolumn{1}{c}{} & \multicolumn{4}{c}{CGA-Wiki (Mistral + SO + G)} & \multicolumn{4}{c}{BNC (Mistral + SO + G)} \\
\cmidrule(r){2-5} \cmidrule(r){6-9}
\# of Generations (L) & Acc. & Prec. & Rec. & F1 & Acc. & Prec. & Rec. & F1 \\
\midrule
1  & 64.2 & 64.7 & 62.7 & 63.7 & 89.2 & 84.9 & 94.4 & 89.4 \\
3  & 64.7 & 65.0 & 63.9 & 64.4 & 89.9 & 86.7 & 93.6 & 90.0 \\
5  & 66.5 & 66.9 & 65.3 & 66.1 & 90.3 & 86.8 & 94.4 & 90.4 \\
7  & 64.9 & 65.6 & 62.9 & 64.2 & 90.3 & 86.2 & 95.2 & 90.5 \\
11 & 65.3 & 66.0 & 63.2 & 64.5 & 90.7 & 86.9 & 95.2 & 90.8 \\
15 & 66.6 & 67.6 & 63.9 & 65.7 & 90.3 & 86.2 & 95.2 & 90.5 \\
\bottomrule
\end{tabular}
\caption{\label{tab:performance_cga_wiki_continuations_ablation} Accuracy, precision, recall, and F1 scores of the Mistral-based classifier on the validation sets of \textbf{CGA-Wiki} and \textbf{BNC}, varying the number of generated future conversations per sample at test time. SO denotes stands for orientation labels, and G stands for generation. Increasing the number of generations per conversation ($L$) from 1 to 5 improves performance on both datasets, but further increases beyond $L=5$ yield diminishing returns. We therefore set $L=5$ in all experiments to balance predictive performance with computational efficiency.}
\end{table*}

To determine $L$, the number of continuations per test sample, we conduct an ablation study on the CGA-Wiki and BNC validation sets, experimenting with $L$ between 1 and 15. 

Intuitively, generating multiple conversation continuations and applying a majority vote over the resulting derailment predictions can reduce the variance introduced by the stochastic nature of LLM sampling, thereby improving prediction accuracy. However, generating too many continuations per conversation can be computationally costly. 

Due to the large number of generations required for this experiment, we adopt vLLM \cite{vllm} to improve generation speed, while the rest of the experiments in the paper use native PyTorch to ensure numerical consistency between training and testing environment.

As shown in Table \ref{tab:performance_cga_wiki_continuations_ablation}, increasing the number of generations $L$ from 1 to 5 leads to 1-2\% improvement in accuracy, along with improvement in precision, recall, and F1. Beyond $L=5$, however, additional gains are marginal. Based on these findings, we set $L=5$ for all experiments in this paper to maintain a reasonable trade-off between predictive accuracy and computational overhead. 

\subsection{Human Evaluation of Social Orientation Labels}
\label{ssec:so_human_eval}

\begin{figure}[h!]
    \centering
    \includegraphics[width=0.8\columnwidth]{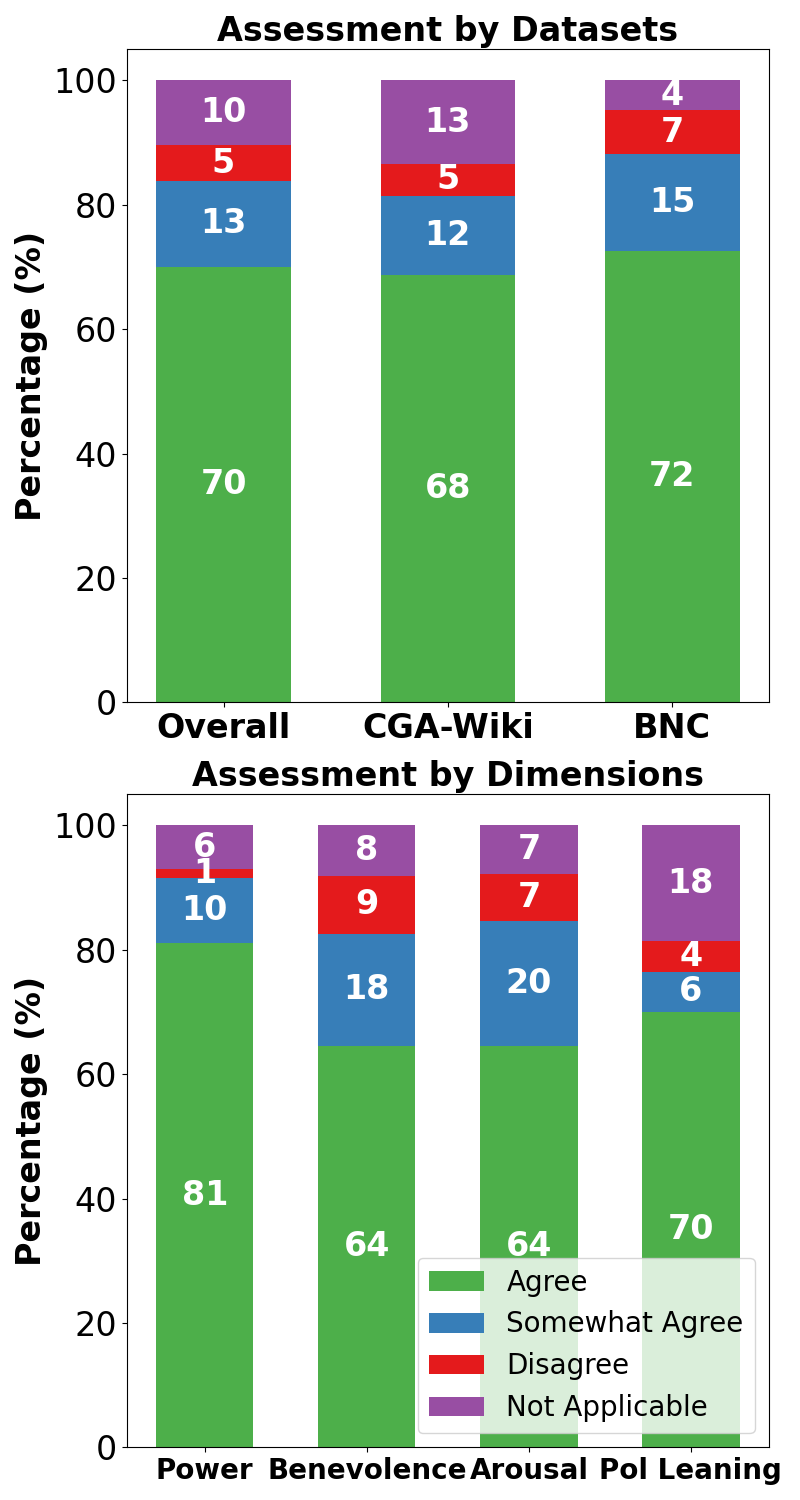}
    \caption{Human evaluation results for the accuracy of GPT-4o-annotated social orientation labels. Overall, the GPT-4o annotations exhibit good quality, with human evaluators agreeing with the predicted labels 70\% of the time.}
    \label{fig:so_human_eval}
\end{figure}

We conducted a human evaluation study to assess the quality of the social orientation labels generated by GPT-4o. We randomly selected 25 conversations from the CGA-Wiki dataset and 25 from the BNC dataset, resulting in a total of 50 conversations. We recruited 6 graduate and undergraduate research assistants from our university as annotators. All annotators consented to the annotation and were not related to this research project otherwise. To estimate inter-annotator agreement (IAA), we assigned 10 out of the 50 conversations to two annotators. Each annotator therefore annotated 10 conversation. We found the Krippendorff's alpha to be 0.186.

\begin{figure*}[!t]
    \vspace*{-13cm} 
    \centering
    \includegraphics[width=\textwidth]{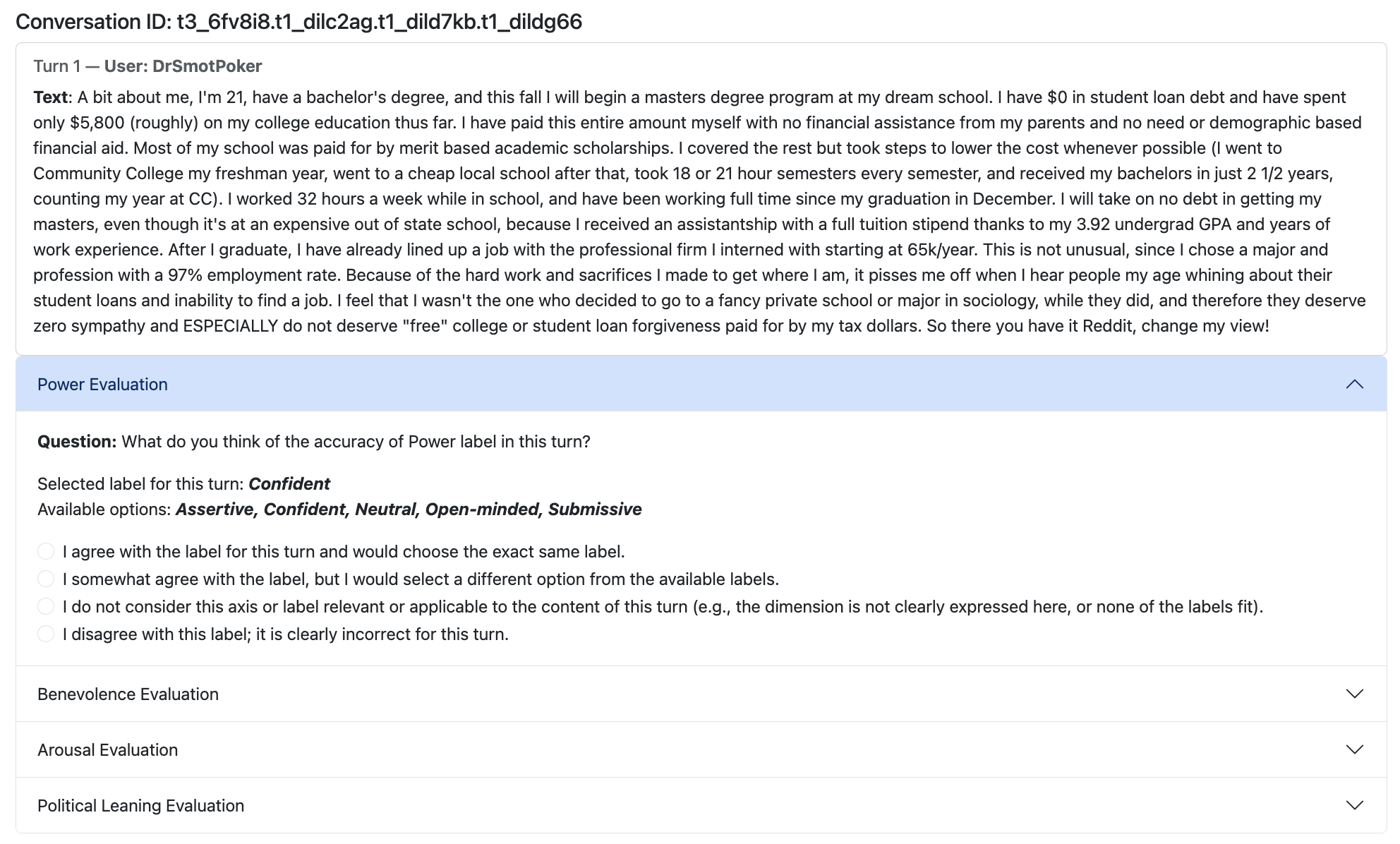}
    \caption{User interface for the human evaluation of social orientation labels. Human annotators were asked to evaluate label quality turn-by-turn and axis-by-axis.}
    \label{fig:human_eval_ui}
\end{figure*}

As shown in Figure \ref{fig:human_eval_ui}, during the evaluation process, annotators assessed the correctness of the GPT-4o-generated social orientation labels on a turn-by-turn, axis-by-axis basis. Annotators assigned each label to one of the following four categories:

\begin{description}[align=left, leftmargin=0pt, style=unboxed] 
\item[Agree:] Agree and would choose the exact same label.

\item[Somewhat Agree:] Somewhat agree but would prefer a different label.

\item[Disagree:] Disagree with the selected label, as it is incorrect.

\item[Not Applicable:] The provided axis or label options are not applicable to this turn.
\end{description}

The evaluation results, shown in Figure \ref{fig:so_human_eval}, demonstrate that GPT-4o produces reasonably accurate social orientation labels. Human annotators fully agreed with GPT-4o's label choices 70\% of the time, while only 15\% of labels fell into the "Disagree" or "Not Applicable" categories. Annotators reported higher agreement for the BNC dataset (72\%) compared to the CGA-Wiki dataset (68\%). Axis-wise analysis revealed that the power axis had the highest agreement (81\%), followed by political leaning (70\%), and benevolence and arousal (both 64\%).

While the human evaluation indicates that GPT-4o achieves reasonably accurate social orientation labels, it is notable that our proposed method's conversation derailment prediction performance (71.4\% for CGA-Wiki and 93.8\% for BNC) exceeds the labeling accuracy (68\% for CGA-Wiki and 72\% for BNC). This discrepancy suggests that the information provided by the social orientation labels might be too noisy to be effectively utilized by the conversation outcome predictor.
\clearpage

\clearpage

\subsection{Model Training and Evaluation Details}
\label{ssec:model_training_evaluation_details}
\begin{table}[H]
\fontsize{9}{9}\selectfont
\renewcommand{\arraystretch}{1.2}
\centering
\begin{tabularx}{\columnwidth}{lX}
\toprule
Hyper-parameter & Value \\
\midrule 
Base Model & mistralai/Mistral-7B-v0.1 \\
Number of Parameters & 7.24 Billion \\
Use LoRA? & True \\
LoRA Rank & 64 \\
LoRA Alpha & 64 \\
LoRA Modules & All Except The Embedding Layer \\
Use LoRA Bias & True \\
Epochs & 3 \\
Max Context Length & 3,072 Tokens \\
Batch Size & 32\\
Optimizer & AdamW \\
LR Schedule & One Cycle Cosine LR with Linear Warmup \\
Max LR & $1 \times 10^{-4}$ \\
Min LR & $2 \times 10^{-5}$ \\
Gradient Clip & 5.0 \\
PyTorch Version & 2.3.1+cu118 \\
HF Transformers Version & 4.42.3 \\
HF PEFT Version & 0.11.1 \\
GPU Model & NVIDIA A100 \\
\bottomrule
\end{tabularx}
\caption{\label{tab:gen-llm-hyper-params} Hyper-parameters for fine-tuning Mistral-7B for conversation generation.}
\end{table}

\begin{table}[H]
\fontsize{9}{9}\selectfont
\renewcommand{\arraystretch}{1.2}
\centering
\begin{tabularx}{\columnwidth}{lX}
\toprule
Hyper-parameter & Value \\
\midrule 
Base Model & mistralai/Mistral-7B-v0.1 \\
Number of Parameters & 7.24 Billion \\
Initial Context Length & 2,048 \\
Max Tokens to Generate & 1,024 \\
Temperature & 1.0 \\
Top P & 0.9 \\
Top K & Not Used \\
Repetition Penalty & 1.05 \\
Generation Batch Size & 8 \\
PyTorch Version & 2.3.1+cu118 \\
HF Transformers Version & 4.42.3 \\
HF PEFT Version & 0.11.1 \\
GPU Model & NVIDIA A100/A6000 \\
\bottomrule
\end{tabularx}
\caption{\label{tab:gen-llm-hyper-params-inference} Hyper-parameters for sampling future conversations from the fine-tuned Mistral-7B model.}
\end{table}

\newpage

\begin{table}[H]
\fontsize{9}{9}\selectfont
\renewcommand{\arraystretch}{1.2}
\centering
\begin{tabularx}{\columnwidth}{lX}
\toprule
Hyper-parameter & Value \\
\midrule 
Base Model & facebook/bart-base \\
Number of Parameters & 139 Million \\
Use LoRA? & False \\
Loss Function & Binary Cross Entropy \\
Epochs & 15 \\
Max Context Length & 1,024 Tokens \\
Batch Size & 32\\
Optimizer & AdamW \\
LR Schedule & One Cycle Cosine LR with Linear Warmup\\
Max LR & $2 \times 10^{-5}$ \\
Min LR & $2 \times 10^{-6}$ \\
Gradient Clip & 5.0 \\
PyTorch Version & 2.3.1+cu118 \\
HF Transformers Version & 4.42.3 \\
HF PEFT Version & 0.11.1 \\
GPU Model & NVIDIA A100 \\
\bottomrule
\end{tabularx}
\caption{\label{tab:bart-classifier-hyper-params} Hyper-parameters for training BART-based conversation classifiers.}
\end{table}

\begin{table}[H]
\fontsize{9}{9}\selectfont
\renewcommand{\arraystretch}{1.2}
\centering
\begin{tabularx}{\columnwidth}{lX}
\toprule
Hyper-parameter & Value \\
\midrule 
Base Model & mistralai/Mistral-7B-v0.1 \\
Number of Parameters & 7.24 Billion \\
Use LoRA? & True \\
LoRA Rank & 64 \\
LoRA Alpha & 64 \\
LoRA Modules & All Except Embedding Layer \\
Use LoRA Bias & True \\
Loss Function & Binary Cross Entropy \\
Epochs & 15 \\
Max Context Length & 2,048 Tokens \\
Batch Size & 32\\
Optimizer & AdamW \\
LR Schedule & One Cycle Cosine LR with Linear Warmup \\
Max LR & $1 \times 10^{-4}$ \\
Min LR & $2 \times 10^{-5}$ \\
Gradient Clip & 5.0 \\
PyTorch Version & 2.3.1+cu118 \\
HF Transformers Version & 4.42.3 \\
HF PEFT Version & 0.11.1 \\
GPU Model & NVIDIA A100 \\
\bottomrule
\end{tabularx}
\caption{\label{tab:mistral-classifier-hyper-params} Hyper-parameters used for training Mistral-based conversation classifiers.}
\end{table}

\clearpage

\subsection{GPT-4o Prompt Template For Social Orientation Annotation}
\label{text:social-orientation-annotation-template} 
\lstset{breaklines=true} 
\begin{lstlisting}
Analyze the communication styles in the specified Wikipedia editor discussions according to four dimensions: power, benevolence, arousal, and progressiveness. Definitions and response options for each dimension are provided below. Begin by reading the first four conversations. For the fifth conversation, annotate every comment according to the dimensions provided, using the same format. Select the most appropriate option from each category for each comment. If a conversation has been partially annotated, only provide annotations for the remaining comments. Provide these annotations directly, without additional explanations or digressions.

Dimensions:

1. Power: This dimension gauges the extent to which an individual seeks to control or assert dominance in a conversation.
- Options: Assertive, Confident, Neutral, Open-minded, Submissive

2. Benevolence: This measures the warmth and positivity of the interactions.
- Options: Confrontational, Dismissive, Neutral, Friendly, Supportive

3. Arousal: This refers to the level of energy and excitement in the comment.
- Options: Energetic, Neutral, Calm

4. Progressiveness: This assesses the political orientation conveyed in the comment.
- Options: Liberal, Neutral, Conservative

In the following conversations drawn from Wikipedia discussion forums, each row corresponds to a turn number, an user name, and a comment made by that user. Provide a social orientation tag for every turn in the input, and do not skip any turns. Closely follow the format in the first four examples, and finish the last sample. Do not provide any explanations.

Conversation 1:
Turn 1: Tryptofish: == Good work! ==
Turn 2: Tryptofish: '''The Admin's Barnstar''' For the apparently thankless task of drafting a suggested closing summary at the RfC/U.
Turn 3: The Wordsmith: Thank you for your kindness. I do make an effort to be even-handed, no matter what people wiki_link about me.
Turn 4: Lar: I was just popping by to offer some words of encouragement. Glad to see Tryp beat me to it. ++: /

Annotations:
Turn 1: Open-minded, Supportive, Energetic, Neutral
Turn 2: Open-minded, Supportive, Energetic, Neutral
Turn 3: Open-minded, Friendly, Neutral, Neutral
Turn 4: Open-minded, Supportive, Energetic, Neutral

Conversation 2:
(Omitted for brevity...)

Social Orientation Tags:
(Omitted for brevity...)

Conversation 3:
(Omitted for brevity...)

Social Orientation Tags:
(Omitted for brevity...)

Conversation 4:
(Omitted for brevity...)

Social Orientation Tags:
(Omitted for brevity...)

Conversation 5:
{Comments to Annotate}

Social Orientation Tags:
\end{lstlisting}


\end{document}